\documentclass[conference]{IEEEtran}
\IEEEoverridecommandlockouts
\usepackage{cite}
\usepackage{amsmath,amssymb,amsfonts}
\usepackage{algorithmic}
\usepackage{graphicx}
\usepackage{textcomp}
\usepackage{gensymb}
\usepackage{url}
\usepackage{hyperref}
\usepackage{subcaption}
\usepackage[numbers]{natbib}
\usepackage[table*]{xcolor}
\usepackage{color, colortbl}

\definecolor{grey}{rgb}{0.7,0.7,0.7}

\def\BibTeX{{\rm B\kern-.05em{\sc i\kern-.025em b}\kern-.08em
    T\kern-.1667em\lower.7ex\hbox{E}\kern-.125emX}}

\newcommand{\emailaddress}[1]{\href{mailto:#1}{\nolinkurl{#1}}}

\begin{document}

\title{RoboMem: Giving Long Term Memory to Robots\\
}

\author{\IEEEauthorblockN{Ifrah Idrees}
\IEEEauthorblockA{\textit{Dept. of Computer Science} \\
\textit{Brown University}\\
Providence, USA \\
\emailaddress{ifrah_idrees@brown.edu}}
\and
\IEEEauthorblockN{Steve P. Reiss}
\IEEEauthorblockA{\textit{Dept. of Computer Science} \\
\textit{Brown University}\\
Providence, USA \\
\emailaddress{spr@cs.brown.edu}}
\and
\IEEEauthorblockN{Stefanie Tellex}
\IEEEauthorblockA{\textit{Dept. of Computer Science} \\
\textit{Brown University}\\
Providence, USA \\
\emailaddress{stefie10@cs.brown.edu}}
}

\maketitle
\begin{abstract}
  
Robots have the potential to improve health monitoring outcomes for the elderly by providing doctors, and caregivers with information about the person's behavior, health activities and their surrounding environment. 
Over the years, less work has been done to enable robots to preserve information for longer periods of time, on the order of months and years of data, and use this contextual information to answer queries. 
Time complexity to process this massive sensor data in a timely fashion, inability to anticipate the future queries in advance and imprecision involved in the results have been the main impediments in making progress in this area. We make a contribution by introducing \textbf{RoboMem}, a query answering system for health-care assistance of elderly over long term; continuous data feeds that intends to overcome the challenges of giving long term memory to robots. The design for our framework preprocesses the sensor data and
stores this preprocessed data into the database. This data is updated in the database by going through successive refinements, improving its accuracy for responding to queries. If data in the database is not enough to answer a query, a small set of relevant frames (also obtained from the database) will be reprocessed to obtain the answer. [Our initial prototype of RoboMem stores 3.5MB of data in the database as compared to 535.8MB of actual video frames and with minimal data in the database it is able to fetch information fundamental to respond to queries in 0.0002 seconds on average].   
\end{abstract}
\section{Introduction} \label{introduction}
One potential use of robots in the personal assistance of the elderly is to answer questions about their health or their surrounding environment. Such questions can span over days, months or even years of data. Getting these queries answered accurately and timely by robots involves the intersection of many disciplines - robotics, computer vision, databases, natural language processing, and human-computer interaction. Although efforts have been made by \citet{Temporal-Grounding} to enable robots to respond to commands  by incorporating the factual knowledge or observations of its workspace for the last few minutes, this has only yielded a short term memory. We intend to go beyond and provide robots the capability to answer queries over long term memory.

Several challenges exist in building an end-to-end system for giving robots long term memory. These challenges include:
\begin{itemize}
    \item There is massive sensor data and this information needs to be stored compactly because of the limited storage.
    \item Initial processing of robot sensor information is not precise enough to answer queries.
    \item The possible queries are not known in advance therefore we cannot determine what information should be stored to answer them. 
    \item Queries should be responded in a timely fashion, reprocessing all original sensor data cannot be afforded for every question
    
    
\end{itemize}
\begin{figure}[t]
    \centering
    \includegraphics[width=0.7\linewidth,height=6.5cm]{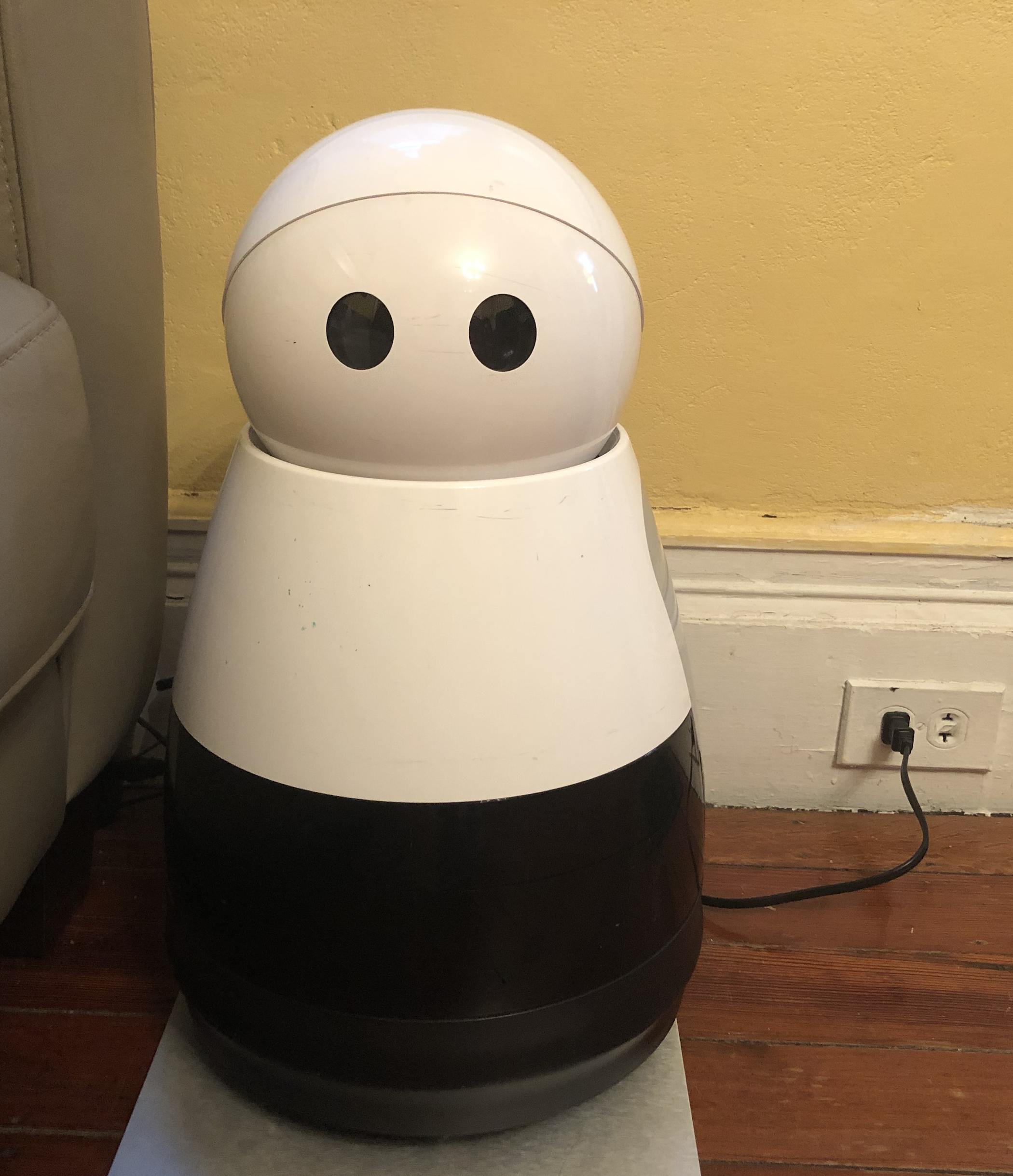}
    \caption{Kuri Robot \citep{richardson_2018}}\label{fig:a}
    \label{fig:Kuri}
\end{figure}

\begin{figure*}[h]
\centering
\includegraphics[width=0.85\textwidth]{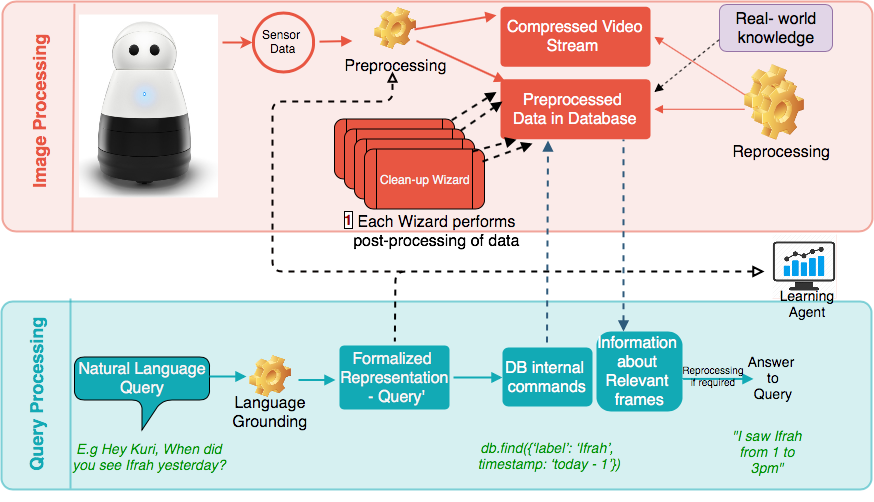}
\caption{Workflow Model of RoboMem}
\label{fig-modules}
\end{figure*}
 
Consequently, we propose a framework; RoboMem, which offers a new approach to robotic memory and perception. RoboMem intends to answer queries that can help assist the care of elderly over periods of days, months, or even years. In particular, we will be providing the social and interactive Kuri robot from  \citet{heykuri_2018} with a long term memory. 

RoboMem intends to overcome the challenges mentioned above with the following design vision:
\begin{itemize}
    \item Formulating a set of categories for queries which are important for the elderly health-care environment.
    \item Proposing an architecture which divides data processing into hierarchical
    phases (pre-processing, post-processing, re-processing).
    \item Successive refinement of the prepossessed data in the database.
    \item Answering queries by reprocessing small subset of original sensor data as needed.
\end{itemize}

\citet{UserStudy} conducted a user study highlighting the top 14 activities important in the daily lives of the elderly. We use it as a ground for constraining the set of queries that will be potentially asked from RoboMem. We have cataloged these queries into categories, and these categories along with examples queries are as follows:
\vspace*{-0.22\baselineskip}
\begin{enumerate}
\item Spatiotemporal object localization in the surrounding of elderly, e.g. - \textit{``Hey Kuri, do you remember where I last placed my daughter's graduation picture?''}
\item Identifying the people elderly has interacted with e.g. - \textit{``Hey Kuri, can you tell me which of my family members visited me last summer?''}
\item Recognition and spatiotemporal localization of activities performed by elderly e.g.
\begin{itemize}
    \item Was the activity performed - \textit{``Hey Kuri, has my dad taken the Montelukast medicine over the past month?''}
    \item How long was the activity performed - \textit{``Hey Kuri, how has my patient's sleeping cycle been (how much has my patient slept) over the last three months?''}
    \item Where was the activity performed -\textit{``Hey Kuri, where did my patient exercise or walk the most over past week?''}
\end{itemize}
\end{enumerate}

Input from a physician working with the elderly was used to assist in developing the queries for our initial prototype, and these queries were personally recommended by him since the elderly nearly always have trouble answering these questions by themselves.

The main contributions of our paper are to enable answering questions about elderly's health and their surrounding robustly and efficiently by proposing the \textit{RoboMem} framework that:
\begin{itemize}
    \item preprocesses the video feed from Kuri in real-time;
    \item organizes the data and stores the prepossessed data in document-oriented database;
    \item performs successive refinement of the data to update data in the database;
    \item uses a subset of preprocessed data along with the query to be directed to original frames that need to be reprocessed for addressing the query for which data is not present in the database.
\end{itemize}
\section{Related Work}
\citet{Temporal-Grounding} worked on extending the space of commands that a robot can understand. However, their system only incorporates factual information from past visual observations and linguistic interactions for around five minutes of data. In their system, increasing the length of the videos increases the context for inference, but it also increases the chances of failures due to errors in perception. We are working on developing a system that performs successive post-processing of the input sensor data as discussed in section-\ref{Section:Framework} to increase the certainty of the stored information and reduce the amount of reprocessing needed to answer the queries. 

In recent years, computer vision has achieved significant success both in terms of accuracy and efficiency for object detection \cite{Chang2011} and face identification \cite{parkhi2015deep}. Researchers have come up with various object detection algorithms such as Mask-RCNNs \cite{MaskRCNN}, RetinaNet \cite{RetinaNet}, and further efforts have been made to make these convolution networks even faster - \cite{FastCNN}, \cite{FasterRCNN}, \cite{RFCNs}. 
Likewise, learning deep video-representations (features) for activity recognition via convolution network has been receiving increasing attention \cite{DAR1}, \cite{DAR2}, \cite{DAR3}, \cite{DAR4}. These deep networks have achieved high recognition performance in a variety of action datasets \cite{DAR6}. RoboMem will be using these state of the art techniques to pre-process data in real time for extracting necessary information from the video sensor data.




\citet{chung2016iros} discusses question answering systems for autonomous mobile robots. Their system stores static and dynamic information of the indoor office environment in a world map and uses that to answer both information acquisition and information retrieval queries, however, their system does not handle questions that span over a period of time and those involving object search. Our framework of RoboMem tries to incorporate both of these functionalities. 

\section{Technical Approach}\label{Section:Framework}
In this section, we discuss the proposed system architecture of RoboMem that will enable robots to have long term memory and answer queries related to the healthcare of elderly and their surroundings. The architecture is also shown in Figure-\ref{fig-modules}.
In our model, RoboMem receives input - video feed and pose information from SLAM navigation which it preprocesses in real time to (i) store preprocessed data into database and (ii) store compressed video stream on external storage 
The preprocessing that we choose for RoboMem is explained in Section-\ref{section-preprocessing}. We envision preprocessing to include object detection on every frame and activity recognition model running on the frames in which a human is detected. Successive refinement of this preprocessed data needs to be performed to achieve higher accuracy and robustness \cite{jones2018recurrent}, \cite{mulit-frames}. RoboMem will accomplish this by passing the preprocessed data through clean-up wizards. The post-processing that will be performed by the clean-up wizards include:
\begin{itemize}
    \item Maintaining probabilities of the 
        \begin{itemize}
        \item Activities recognized and performed by the elderly 
        \item People and objects involved in the activity
        \item Location of where the activity was performed
        \end{itemize}
    \item Updating the probability distribution of various features such as the location of objects over different frames over time;
\end{itemize}
For updating probabilities, these clean-up wizards shown in Figure-\ref{fig-modules} will process combined data from multiple frames. This additional information gained will help increase the certainty of the data stored. 
To handle the case when sufficient data is not present in the database at the time when the query is asked, RoboMem will include a module for reprocessing the relevant video stream frames. A real-life example of this can be when elderly inquires from RoboMem about \textit{``Hey Kuri, can you tell me the days when my grandson was wearing red T-shirt?''} and RoboMem's preprocessed data does not include the color of the T-shirts stored and will just have information of when the grandson was present. To handle this situation, RoboMem will fetch relevant frames that include the grandson and then reprocess frames to extract the color of the T-shirt.

A means of accessing long-term memory of RoboMem is query processing. The eventual goal is to enable RoboMem to understand natural language queries by all elderly, caregivers, and the doctors. This includes translation of the natural language query into an intermediate form - a formal query representation: $Query'$ (Figure-\ref{fig-modules}) which can then be converted into internal DB commands using a method similar to \cite{Dbpal}.
These DB commands will then be used to fetch the attributes with the highest probability which will be used to create a response to the query in natural language. As mentioned previously, these probabilities are stored and being updated in the database. Technical challenges involving language grounding of the queries is discussed in the  section-\ref{section-TechnicalChallenges}.

RoboMem will also be exploring ways to store real-world knowledge since it will be integrating static information about the real world, elderly's life and their surrounding to answer some of the queries. For example, for RoboMem to understand queries like a doctor asking from RoboMem: ``Hey Kuri, has my patient taken her medicines this week?'', RoboMem needs the information of what medicines have been prescribed to the elderly to be able to ground queries and then respond to them accordingly. 
\citet{inproceedings} proposes a multi-level ontology model for storing real-world knowledge. Different information about the environment ({\em e.g., }texture of objects, activities associated with an object) is saved in different ontology layers and answers are obtained using goal based query reasoning while \citet{4059204} represents the environment as a hierarchical graph modeled using a tour-guide like interaction between human(guide) and the robot. \citet{chung2016iros} stores a static 2-D map of the real world, which is subdivided into regions and offices with information of the person to whom the office is assigned.

\section{Technical Challenges} \label{section-TechnicalChallenges}
In this section, we describe various validity and scalability issues that need to be addressed before RoboMem can robustly and efficiently answer queries by doctors, caregivers about the elderly's healthcare, and their surrounding. These challenges are as follows:

\begin{figure}
\centering
\includegraphics[width=1\linewidth]{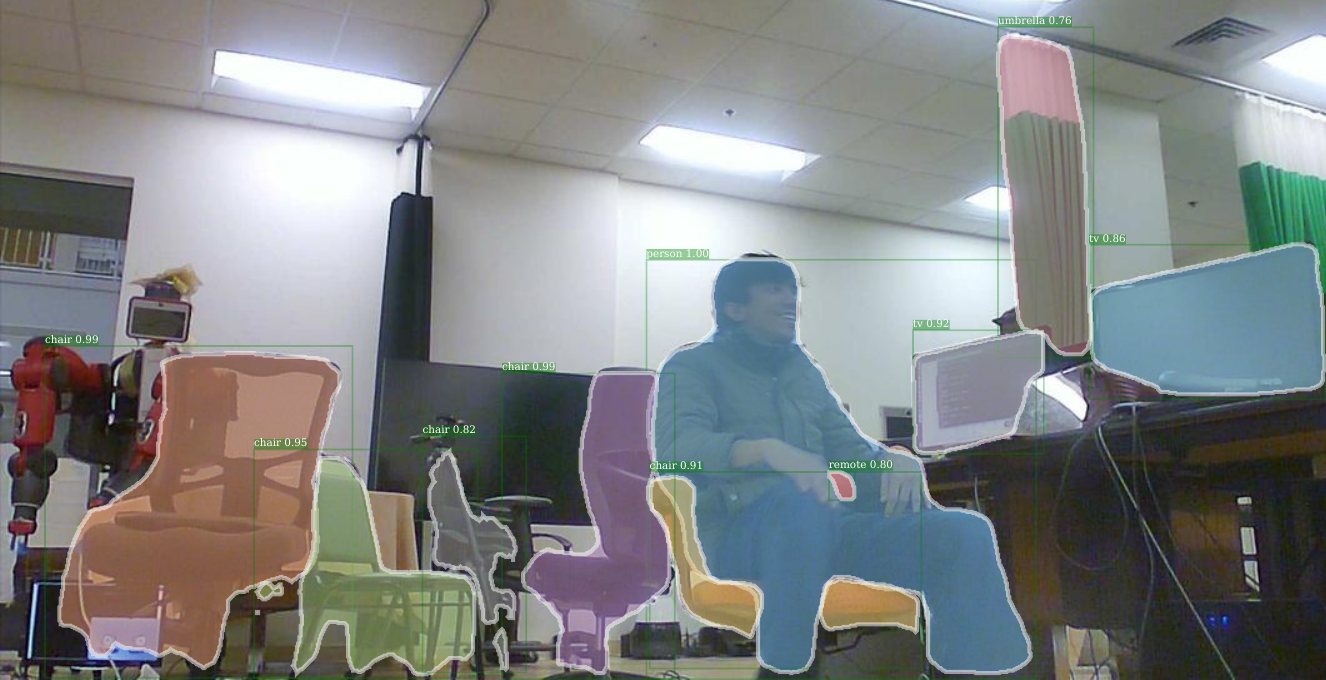}
\caption{Segmented objects in Kuri's image frame}
\label{fig:KuriImageFrame}
\end{figure}

\subsection{Massive Sensor Data}\label{section-preprocessing}
RoboMem receives continuous raw video feed data from sensors as input. If preprocessing of this data is not performed in real-time, frames yet to be processed will continually accumulate and in the worst case they can gather to such an extent that even an overnight preprocessing of this video feed might not help to add relevant information in the database before a related query is asked leading to time-delays in answering them. Further, at the time when a query is asked, RoboMem cannot reprocess all of the sensor data when asked a query since it will hinder the process of answering the queries in a timely fashion. Therefore, RoboMem needs to be designed such that it is able to preprocess sensor data in real-time and store basic information in the database that can be refined in the background and is enough to answer queries in real-time or, when/if required, is able to restrict the number of frames needed to be reprocessed for responding to query. 

We tackle this challenge in the initial study and evaluate the feasibility of our proposed system by developing a prototype that: 
\begin{itemize}
    \item Extracts information from the sensor data that is fundamental to answering queries.
    \item Validates that preprocessing of the video can be done in real-time;
\end{itemize}

Our prototype is described  and evaluated in section-\ref{sec-DescriptionFeasibilityStudy}. Right now preprocessing is done for every frame, but RoboMem will later have to fine-tune variables as to whether processing should be done for every frame to speed up pre-processing. Further, we need to explore if this level of object detection by Detectron will be appropriate for the queries.
We also need to note that stacking object detection and activity recognition will also yield real-time performance, since state-of-art activity detectors can process more frames per second than Kuri's current frame rate which is 6 frames per second \cite{AR7Wang}.
\begin{table*}{}
\centering
 \begin{tabular}{||p{5.5cm} | p{5cm} |p{1cm} | p{5cm}||} 
 \hline
 Query Types & Query Examples & DB Interaction & DB Command\\ [0.5ex] 
 \hline\hline
 Spatio-temporal object localization & {\em Hey Kuri, where did you last see Ifrah?}  & Yes & 
 
 db.find{\normalsize(}\(\{'label'\colon'ifrah'\}\){\normalsize)}.limit{\normalsize(}1{\normalsize)}
 .sort{\normalsize()}{\normalsize[}\('location\_of\_object'\){\normalsize]}
\\
 \hline
 Identifying the people elderly has interacted with  & {\em Hey Kuri, Did Steve visited me yesterday?} & Yes & 
{\em db.find\(\{"label"\colon "Steve",  'timeStamp': new Date(current\_date-1,time:00:00am), \$lt: new Date(currentdate)\}\)}
\\ 
 \hline
 Was the activity performed?  & \textit{Hey Kuri, has my dad taken the Montelukast medicine over the past month?} & No & \cellcolor{grey}\\
 \hline
 How long was the activity performed  & \textit{Hey Kuri, how much has my patient slept over the last 3 months?} & No & \cellcolor{grey} \\
 \hline
 Where was the activity performed & \textit{Hey Kuri, where did my patient exercise or walk the most over past week?} & No & \cellcolor{grey} \\ [1ex] 
 \hline
 \end{tabular}
 \caption{DB commands for queries handled by intial prototype}
 \label{tab:1}
\end{table*}


\subsection{Memory Representation}
In our design, both the compressed video stream and data in the database needs to be stored. Although cloud storage can be used to solve the problem of storing compressed video stream, but one cannot merely assume infinite memory. Therefore, with regards to storage of video stream, we can work on saving summaries of really old data. In this regard, we can extend the model of \cite{LTM1},\cite{LTM2} to our context. \citet{LTM1} applies the migration of memory items from short to medium and long term memory. Their model is demonstrated to work on simple situations, developing memory for 4 blocks on the table, we intend to use their work as a foundation and apply on a complex scenario where the environment is dynamic and objects to be detected are not limited.

Secondly, with respect to the database design RoboMem cannot store every information of every object detected in every scene. The database needs to be designed such that it stores basic data and can return a small set of frames which can be processed to derive an answer for the query. The structure of the database that we chose for the initial prototype is described and justified in section-\ref{section-dbdesign}. Our construction of database collection allows us to answer queries related to spatiotemporal localization of objects in the elderly's surrounding and identifying people elderly has interacted with. In the future, we will explore, should data be stored for each frame or only for the keyframes?

\subsection{Post-processing Layer Design}
The information extracted from pre-processing is not going to be precise enough to answer queries. Therefore, RoboMem needs to have a post-processing layer to increase the certainty of data in the database. Clean-up wizards need to maintain consistency while performing this post-processing, as new sensor data is received it needs to update the probabilistic location of the detected objects or humans , {\em e.g.,} the location of the elderly in two frames could be different. Clean-up wizard needs to identify that the person in both the frames is the same elderly and update the information accordingly. For future work, we will be exploring which design of the post-processing layer leads to most efficient look-up for queries that it has not seen before.

\subsection{Cost of Reprocessing Data}
The ability of RoboMem to have long-term memory and respond to queries about the elderly's health relies heavily on appropriate pre-processing as reanalysis of prior image data in the general case is likely to be very expensive. Further, the possibility of queries that can be asked from RoboMem regarding health-care monitoring of elderly even with the constraints of categories mentioned in Section-\ref{introduction} is vast and therefore there will be situations where enough information is not in the database to adequately answer the query leading to the expensive operation of reprocessing the compressed video feed. As discussed above, an architecture that guides to relevant sample frames for reprocessing as opposed to reprocessing the whole video feed of years or even hours of data will help ease this challenge. This can include identifying which objects are important to consider, adding different type of image processing in the pre-processing queue and setting a hierarchy of priorities between them accordingly. Along with this, as an extension to our prototype, we believe that as future work a learning agent needs to be part of RoboMem that will learn from the past queries to adjust the pre-processing accordingly. 





\subsection{Natural Language Grounding of Query}
As discussed in \cite{Dbpal}, a RNN can be used to for the translation of natural language query to MongoDB commands, but this will require manually curated training corpus of natural language sentence - MongoDB commands pair. Even after the translation, one might have to perform post-processing such as DB command correction(finding time-spans for periods like \textit{``Past month''}), or handling of compositional DB commands like aggregations. Further, once RoboMem fetches response for a query related to the elderly, the response will comprise of multiple frames with different associated probabilities. In the future work, we will be exploring ways of grounding natural language queries to fetch responses from the database.

\section{Future Work}
Currently, our prototype processes information in marginal real time for 37 minutes of video as discussed in section-\ref{sec-Quant}. However, as the amount of data increases, improvements need to be made in processing power and techniques. As a part of future work, we intend to address all the challenges mentioned in section-\ref{section-TechnicalChallenges} and build an end-to-end system. We hope to return a small set of relevant frames for visual processing in the next eight months. Next, we will work on grounding natural language query instead of working with intermediate queries for our system for another year or so.
 
\section{Initial Feasibility Study} \label{sec-DescriptionFeasibilityStudy}

\subsection{Extraction of Data}\label{sec-desstart}
For our prototype of RoboMem, we allow the Kuri Robot \citep{richardson_2018} from Mayfield Robotics to make observations and collect video in an indoor environment setting. We collect 37 minutes of video data with 13320 image frames.

Kuri has 2 RGB-D cameras that capture video feed with a frame rate of 6FPS. For our purposes we are using image stream from the left eye camera. The default size of the images is 1920*1080 pixels, but for speeding up the preprocessing, we resize the images to 1067*600. We also extract Kuri's pose estimate -- x,y,z -- in meters and roll, pitch, yaw of the camera in degrees. We are extracting this sensor data for identifying objects and estimating their location. 
\subsection{Prototype's preprocessing sequence}
Our prototype focuses on extracting the basic data needed to enable further post-processing or reprocessing to answer two types of queries that can be asked from RoboMem by caregivers, doctors or elderly as discussed in section-\ref{introduction} -- Spatiotemporal localization of objects in the surrounding of elderly and identifying the people elderly has interacted with. Our prototype performs real-time object detection by using Facebook's AI Research software system - Detectron \cite{Detectron2018}. Detectron includes implementation of multiple object detection algorithms, and for our implementation of RoboMem we use an end-to-end trained Mask R-CNN model with a ResNet-101-FPN backbone from a model zoo trained by \citet{Detectron2018}. Once the objects are detected in frame $f$, the objects that are humans are labeled manually. Figure-\ref{fig:KuriImageFrame} shows the segmentation of objects that we achieve for a sample image frame captured in the 37 minutes of video. 

The reason we do this kind of preprocessing is to store minimal information in DB for answering different query categories, as mentioned in section-\ref{introduction}, regarding elderly health-care. For each of those categories, we need to perform the fundamental operation of detecting objects that are of primary focus in the query. Take, for example, the lost object identification query by the elderly: \textit{“Hey Kuri, Do you remember where I last placed the Television's remote?”} For this query, at the basic level, RoboMem  needs to detect all the video frames that contain the remote that our prototype is able to detect. This detection is done as a preprocessing step. Another example query for identifying important people can be \textit{"Hey Kuri, can you tell me if Steve visited me last summer?"} This will require RoboMem to detect frames captured within a time frame(last summer) that contain a person labeled as Steve in the database. Through object detection and the extraction of time and robot pose, our prototype can provide us information that is fundamental for answering these queries. RoboMem will later have to fine-tune variables as to whether pre-processing should be done for every frame to speed up pre-processing. 

\subsection{Database Design}\label{section-dbdesign}
For queries related to spatiotemporal localization of objects and identifying people elderly has interacted with - the fields that our prototype integrates in the database collection are: 
\begin{itemize}
    \item Image frame number say $f$;
    \item Objects detected in the frame $f$;
    \item Object labels for objects detected e.g., name of the human;
    \item Robot's pose - x,y,z,roll,pitch,yaw - at frame $f$;
    \item Timestamp when Kuri captured frame $f$;
    \item An estimate of the location - $x$, $y$ coordinates of the objects detected. We assume that the objects detected in the image frame are at the center of a circle within the radius of 2m. Our future work includes a more accurate localization of objects.
\end{itemize}

The reason behind choosing such a design structure for the initial database collection is that in the case of lost object identification, our prototype with such a design can return key information like the latest timestamp for the frame in which it saw the object. Likewise, for the identification of important people, it returns whether it has seen that specified person - the person named Steve - in the database. RoboMem is able to respond to such queries, without having to reprocess the whole video feed by having a structure that stores minimal information in the database.

We implement our prototype for RoboMem using python API for MongoDB - pyMongo. The choice to use MongoDB over the other databases was because of its flexibility to augment fields in the collections, which will be useful because of our multi-tier processing.  Furthermore, in the future we intend to expand our framework to cloud databases, and MongoDB provides additional benefits of supporting distributed systems, allowing us to handle our application at scale \cite{mongodb}.

\subsection{Quantitative Evaluation of prototype}\label{sec-Quant}
The image frames are sent from Kuri to a standalone desktop for image processing. The time taken to transfer these images can be ignored because of the high speed network connection. Detectron has an inference time of 0.143 seconds per image, on average. The total time to organize preprocessed data for 13320 frames and inserting it in the mongoDB requires 59.382 seconds while the average time to insert a document with 6 fields mentioned in section \ref{section-dbdesign} in the database was 0.0041 seconds. Thus, with more sophisticated preprocessing, e.g., extracting depth of the objects, we want to be able to insert preprocessed data in the database with improved performance in real-time.

The size of the actual video and metadata (robot pose and timestamps) for the jpeg frames was 535.8MB while the size of 37 minutes of video feed encoded in mp4 was 79.5 MB, compared to the database size of only 3.5MB. These numbers not only show the amount of space we save after preprocessing the data but also that storing the feed as a video rather than individual images is much more scalable.

Table-\ref{tab:1} shows the DB commands that we use in our prototype to fetch data for the queries that we intend to address, as mentioned in Section-\ref{introduction}. 

\section{Conclusion}
We design a query framework for robots to enable monitoring health-care of elderly and their surrounding environment by responding to queries over long, continuous feed of sensor data. We discuss the technical challenges in enabling robots to develop long-term memory and draft an architecture to overcome these challenges by performing i) real-time preprocessing of massive video sensor data (ii) successive refinement of data in database performed by clean-up wizards, and augmentation of this data into the database; (iii) reprocessing of data if enough information is not in the database; (iv) probabilistic language grounding for query processing; (v) learning from past queries to adjust the prepossessing queue and/or structure of MongoDB collections. We also conduct an initial study to show the feasibility of our proposed system. We believe that with these aforementioned components, RoboMem will be able to overcome the limitations involved in giving long term memory to robots and help provide caregivers and the elderly with information about the elderly and their surrounding.

{\scriptsize
\bibliographystyle{plainnat}
\bibliography{mybib}}

\end{document}